\pdfoutput=1  
\documentclass[10pt,twocolumn,letterpaper]{article}

\usepackage[pagenumbers]{cvpr}   

\providecommand{\natexlab}[1]{}

\newcommand{\best}[1]{\textbf{#1}}
\newcommand{\second}[1]{\underline{#1}}

\definecolor{cvprblue}{rgb}{0.21,0.49,0.74}
\usepackage[breaklinks,colorlinks,allcolors=cvprblue]{hyperref}

\def\confName{CVPR}
\def\confYear{2027}

\title{ForeseeDrive: Future-Token Conditioning for Anticipatory Generative Driving}

\author{%
Shaokai Wu$^{1,2}$, \quad Xunchu Zhou$^{1,2}$, \quad Pinzhi Wang$^{2}$, \quad Bo Dong$^{2}$\\[3pt]
Yuting Hu$^{1}$, \quad Yanbiao Ji$^{1}$, \quad Qiuchang Li$^{1}$, \quad Xiao Cui$^{2}$\\[3pt]
Qi Kong$^{2}$, \quad Hongtao Lu$^{1}$, \quad Yue Ding$^{1}$\\[10pt]
{\small $^{1}$Shanghai Jiao Tong University \qquad $^{2}$Zelos Technology}}

\begin{document}
\maketitle

\begin{abstract}
Generative driving planners can represent diverse ego futures, but a downstream scorer can choose only among sampled candidates; it cannot recover an absent maneuver. We present ForeseeDrive, which turns future scene evolution into a deployable interface for shaping candidate generation. Rather than reconstructing dense future states or scoring a fixed menu, ForeseeDrive derives query-indexed tokens from logged actor motion, occupancy evolution, route interactions, and TTC margins, and predicts the same token space from present context at deployment. The tokens condition trajectory geometry before candidates materialize. A zero-token-gated residual path and candidate-preservation loss limit unnecessary drift, while a TTC-margin guard targets brittle safe-to-unsafe boundary crossings. Under their respective benchmark protocols, ForeseeDrive reaches 89.2 PDMS on NAVSIM navtest, 0.41\,m average L2 and 0.05\% collision rate on nuScenes validation, and a 76.50 driving score on Bench2Drive. Controlled DiffusionDrive/NAVSIM diagnostics show an exact zero-token no-op, a small aligned-over-shuffled selected-PDM advantage under preservation, mixed oracle-best quality, and localized TTC-boundary failures. Together, the benchmark results and bounded mechanism tests support future-conditioned generation as a practical interface for anticipation. Code is included in the supplementary materials.

\end{abstract}

\section{Introduction}
\label{sec:intro}

\begin{figure}[t]
  \centering
  \includegraphics[width=\linewidth]{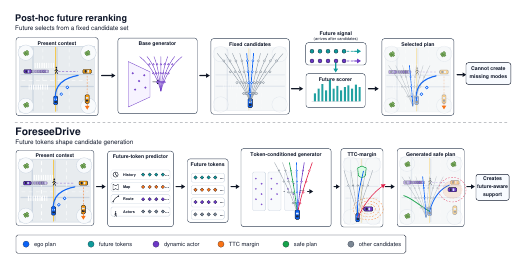}
  \caption{Top: fixed-candidate future scoring acts after a base generator produces candidates, so it cannot create an absent future-aware mode. Bottom: ForeseeDrive predicts compact future tokens from present context and conditions the generator before materialization; a TTC-margin guard rejects brittle near-contact transitions while candidate preservation limits drift.}
  \label{fig:1}
\end{figure}

Autonomous driving planning maps recent observations, map context, and route goals into safe, progressive ego motion~\citep{caesar2020nuscenes,sun2020waymo,chang2019argoverse,wilson2023argoverse2,caesar2022nuplan,dauner2024navsim}. Recent end-to-end and generative planners reduce brittleness by representing multiple ego futures rather than committing to one trajectory~\citep{chitta2022transfuser,hu2023uniad,jiang2023avad,liao2025diffusiondrive}. Yet a generative planner can still fail even when the appropriate behavior is clear in hindsight. The challenge is not only uncertain interaction: the planner must instantiate ego modes before those interactions fully unfold. Once a candidate set is materialized, a downstream scorer can only reorder sampled trajectories; it cannot recover a yielding, braking, nudging, or passing maneuver absent from that set. As Figure~\ref{fig:1} illustrates, anticipation can therefore fail as a support-allocation problem before it becomes a ranking problem.

This view exposes a mismatch between future prediction and trajectory generation. Existing future-centric methods demonstrate the value of actor forecasting, occupancy/BEV prediction, and future-aware planning~\citep{gupta2018socialgan,jiang2023bmotiondiffuser,hu2021fiery,mahjourian2022occupancy,zhang2022beverse,zheng2023occworld,zhang2025seerdrive}. A dense rollout, however, describes the scene broadly, whereas a generator needs compact future evidence that changes which ego modes receive probability mass. Conversely, future-aware scoring introduces anticipation only after the candidate menu is fixed and remains bounded by sampled support. What is missing is a deployable, planner-facing future representation that acts before candidate materialization.

We propose ForeseeDrive to provide this interface. Its central idea is to distill future scene evolution into compact tokens that answer action-relevant planning queries and enter the generator before sampling. During training, logged actor motion, occupancy changes, route interactions, and time-to-collision (TTC) margins supervise the token space; at deployment, a predictor estimates the same variables using only present observation history, map, and route. The predicted tokens can reshape trajectory geometry and mode allocation rather than only the final scores of a fixed set.

The resulting representation is queryable and action-facing: tokens answer local planning questions rather than reconstruct every actor state or BEV cell. They capture whether an actor is becoming route-relevant, whether a reachable corridor will be contested or cleared, and whether an interaction lies near a TTC-sensitive boundary. This makes the representation lower-dimensional than a dense rollout, richer than a scalar risk score, and better aligned with generative planning decisions.

Generation-time future conditioning creates a second tension: the signal must be expressive enough to move candidate modes, yet restrained enough not to corrupt strong candidates from a pretrained planner. ForeseeDrive resolves this tension with an architectural zero-token no-op, candidate preservation, and a margin-aware TTC guard. The preservation loss limits nonzero-token drift when future evidence is weak or irrelevant. The guard targets narrow safe-to-unsafe boundary crossings rather than imposing a broad clearance preference, which would also penalize controlled interactions~\citep{shalev2018safe,dauner2023parting,dauner2024navsim}.

Experiments evaluate both benchmark performance and the proposed mechanism. On NAVSIM navtest, ForeseeDrive reaches 89.2 PDMS and 83.5 EP; on nuScenes validation, it attains 0.41\,m average L2 and 0.05\% collision rate; and on Bench2Drive, it reaches a 76.50 driving score and 53.06 success rate. Same-backbone diagnostics then separate generation from selection: zero tokens exactly preserve the base candidates, aligned tokens modestly outperform shuffled tokens on selected PDM under preservation, and TTC replay localizes boundary-crossing failures. Oracle-best quality remains mixed, so we interpret these diagnostics as evidence for controllable support allocation, not universal support expansion.

Our contributions are threefold. First, we formulate anticipation in generative driving as a support-allocation problem and expose the structural limit of fixed-candidate future-aware scoring. Second, we introduce a deployable future-token interface that distills logged future evolution into queryable, action-facing variables and injects them before candidate materialization. Third, we develop a restrained integration strategy with exact zero-token preservation, base-candidate regularization, and a TTC-margin guard; we evaluate it across three benchmarks and use controlled diagnostics to separate fixed-menu scoring, mode selection, and generation-time conditioning.

\section{Related Work}
\label{sec:related}

\paragraph{Generative planning for autonomous driving.} Generative planners model a distribution over ego futures and materialize multiple trajectories for downstream selection. Diffusion-based planners use iterative denoising and guidance to represent multimodal behavior in prediction, planning, and control~\citep{ho2020ddpm,song2021score,song2021ddim,janner2022diffusion,chi2024diffusionpolicy,jiang2023bmotiondiffuser,zheng2025flexible,liao2025diffusiondrive}; generative end-to-end methods also model ego and traffic evolution in latent or tokenized spaces~\citep{zheng2024genad}. ForeseeDrive complements these generators with supervision that generic guidance does not specify: its query-indexed conditions have logged-future semantics, are predicted online, and enter a zero-token-gated geometry path before selection.

\paragraph{Future prediction and driving world models.} Motion forecasting models socially plausible trajectories, map-conditioned goals, joint interactions, and multi-agent futures~\citep{gupta2018socialgan,chai2019multipath,salzmann2021trajectron,gao2020vectornet,liang2020lanegraph,zhao2020tnt,ngiam2022scene,shi2023mtr,nayakanti2022wayformer,seff2023motionlm}; occupancy and BEV forecasting represent evolving free space or world state~\citep{hu2021fiery,mahjourian2022occupancy,zhang2022beverse,zheng2023occworld}. Driving world models and future-aware planners further use imagined visual, BEV, or occupancy futures to guide planning or trajectory evaluation~\citep{wang2024drivingfuture,wang2024drivedreamer,zhang2025seerdrive,li2025wote}. Our distinction from this line is the planner interface and its timing: ForeseeDrive compresses future evidence into action-facing queries and injects it before finite candidate materialization, rather than requiring a dense rollout or limiting its use to trajectory evaluation.

\paragraph{Structured planning interfaces.} End-to-end driving systems often improve planning by exposing structured intermediate representations, including spatial-temporal BEV features, vectorized scene encodings, object-level planning tokens, interpretable safety cues, and trajectory-guided controls~\citep{hu2022stp3,chitta2022transfuser,hu2023uniad,jiang2023avad,renz2023plant,shao2023safety,wu2022tcp}. These interfaces reduce the burden on a monolithic policy and make planning more controllable. However, most of them primarily encode the current scene, task context, or general safety constraints. ForeseeDrive focuses on a different interface: future-conditioned variables that specify where scene evolution should expand, suppress, or protect ego trajectory modes.

\paragraph{Learning from privileged future information.} Logged driving data contains future scene evolution that is unavailable at deployment, making the problem naturally related to learning using privileged information and teacher-student distillation~\citep{lopezpaz2016privileged,hinton2015distilling}. ForeseeDrive follows this train-deploy separation, but the role of privileged information differs from generic feature distillation. The teacher does not simply provide a stronger representation or a future-aware score; it defines action-facing token targets that the student predicts online and that condition generation before candidates are sampled.

\paragraph{Candidate selection and safety-aware planning.} Candidate scoring and reranking are effective when a planner already generates a sufficiently diverse set of alternatives. However, selection is structurally limited by the sampled support: a scorer cannot recover a yielding, braking, or nudging maneuver that was never generated. Safety-aware planning and benchmark metrics further expose this issue because small trajectory changes near collision or TTC boundaries can produce discontinuous metric changes~\citep{shalev2018safe,dauner2023parting,dauner2024navsim}.

ForeseeDrive therefore combines generation-time future conditioning with candidate preservation and a TTC-margin guard, targeting support allocation and safety-boundary control.

\section{Problem Formulation}
\label{sec:problem}

Let $x_{t} = (\mathcal{H}_{t},\mathcal{M}_{t},\mathcal{R}_{t})$ denote the deployable planning context at time $t$, where $\mathcal{H}_t$ contains recent ego state, sensor history, and tracked actors $\mathcal{A}_{t}$, $\mathcal{M}_{t}$ is the local map, and $\mathcal{R}_t$ is the route command, following common map- and route-conditioned planning formulations~\citep{caesar2022nuplan,dauner2024navsim,chitta2022transfuser,hu2023uniad,jiang2023avad}. The ego plan is a trajectory $\tau = \{(x_s, y_s, \theta_s)\}_{s=1}^T$ over horizon $T$. A base generative planner represents
\begin{equation}
p_{\theta}(\tau \mid x_{t}) = p_{\theta}(\tau \mid \mathcal{H}_{t}, \mathcal{M}_{t}, \mathcal{R}_{t}),
\label{eq:1}
\end{equation}
and materializes a finite candidate set $\mathcal{C} = \{\tau^{(k)}\}_{k = 1}^{K}$ for downstream scoring or selection.

Logged training data additionally contains future scene evolution $\mathcal{F}_{t:t + T}$, including actor motion, occupancy changes, and route-relative interactions. This information is privileged: it may define training targets and diagnostics, but it is not an online input at deployment~\citep{hinton2015distilling,lopezpaz2016privileged}. We use it to define a teacher future-token space
\begin{equation}
\mathbf {Z} ^ {f} = g _ {\phi} (\mathcal {F} _ {t: t + T}, x _ {t}),
\label{eq:2}
\end{equation}
and learn a deployable predictor
\begin{equation}
\hat {\mathbf {Z}} ^ {f} = h _ {\psi} (x _ {t}),
\label{eq:3}
\end{equation}
where $\hat{\mathbf{Z}}^f$ must be estimated only from present observation history, map, and route. The tokens summarize future evidence relevant to ego action rather than reconstructing the full future scene.

The central distinction is where future information enters the planning pipeline. A post-hoc future-aware selector, applied only after candidate generation, first samples candidates from the base planner,
\begin{equation}
\mathcal {C} _ {0} \sim p _ {\theta} (\cdot \mid x _ {t}),
\label{eq:4}
\end{equation}
and then chooses a trajectory using a score that may depend on predicted future tokens:
\begin{equation}
\tau^ {\star} = \arg \max _ {\tau \in \mathcal {C} _ {0}} s (\tau , x _ {t}, \hat {\mathbf {Z}} ^ {f}).
\label{eq:5}
\end{equation}
This selector can reorder trajectories inside the sampled candidate set, but it cannot recover a maneuver that is absent from $\mathcal{C}_{0}$.

In contrast, future-conditioned generation injects predicted tokens before candidate materialization:
\begin{equation}
\mathcal {C} _ {f} \sim p _ {\theta} (\cdot \mid x _ {t}, \hat {\mathbf {Z}} ^ {f}).
\label{eq:6}
\end{equation}
Thus, future information can change not only candidate ranking but also trajectory geometry, mode allocation, and the realized candidate set. This is the setting studied here: privileged future supervision defines action-facing targets during training, a deployable predictor estimates them online, and the resulting tokens condition generation before the candidate set is fixed.

To make this distinction precise, we use empirical candidate support to mean the trajectory modes represented in the realized finite set $\mathcal{C}$. A selector can only change their ordering, whereas a future-conditioned generator can change the distribution that produces them. The goal of ForeseeDrive is therefore not merely to score trajectories with future evidence, but to allocate candidate support toward maneuvers whose value becomes apparent through anticipated interaction.

\section{Methodology: ForeseeDrive}
\label{sec:method}

\begin{figure}[t]
  \centering
  \includegraphics[width=\linewidth]{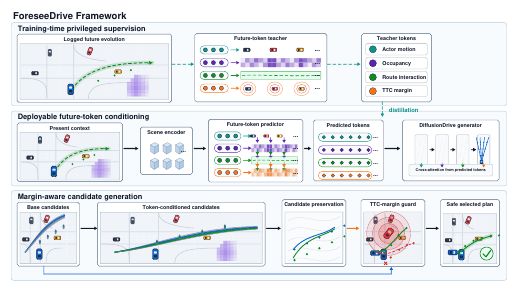}
  \caption{ForeseeDrive framework. Top: logged future evolution defines teacher tokens for actor motion, occupancy, route interaction, and TTC margin. Middle: a deployable predictor estimates matching tokens from present context and conditions the base generator before sampling. Bottom: candidate preservation and a TTC-margin guard constrain the token-conditioned candidates before standard selection.}
  \label{fig:2}
\end{figure}

\subsection{Overview}

We now present ForeseeDrive, a future-token conditioning interface for generative autonomous driving. Given a base planner $p_{\theta}(\tau \mid x_t)$, ForeseeDrive adds anticipation without changing its deployable input space. The method has three stages. First, logged future evolution is converted into query-indexed teacher tokens $\mathbf{Z}^{f}=g_{\phi}(\mathcal{F}_{t:t+T},x_{t})$ during training. Second, a deployable predictor estimates the same token set from online context, $\hat{\mathbf{Z}}^{f}=h_{\psi}(x_{t})$. Third, these predictions condition the generator before candidate materialization, producing $\mathcal{C}_f\sim p_\theta (\cdot \mid x_t,\hat{\mathbf{Z}}^f)$. Candidate preservation and a TTC-margin guard constrain token-induced changes, allowing support to move when predicted future evidence is decision-relevant while protecting strong base candidates.

Figure~\ref{fig:2} summarizes the data flow. Logged future is never exposed at deployment: it defines teacher targets during training, while inference uses predicted tokens from present context. The same interface is attached to the base planner used by each benchmark protocol; the controlled DiffusionDrive/NAVSIM diagnostic is the candidate-level instantiation because it exposes regenerated candidate sets and exact zero-token controls.

\subsection{Query-indexed future tokens}

During training, we construct teacher tokens from logged future evolution using deterministic query-indexed targets. The query set is
\begin{equation}
\mathcal {Q} = \mathcal {Q} _ {a} \cup \mathcal {Q} _ {r} \cup \mathcal {Q} _ {o} \cup \mathcal {Q} _ {m},
\label{eq:7}
\end{equation}
where $\mathcal{Q}_a,\mathcal{Q}_r,\mathcal{Q}_o,$ and $\mathcal{Q}_{m}$ are actor, route, occupancy, and margin queries. Actor queries are the top-$N_a$ tracked actors ranked by route relevance and distance to the ego corridor. Route queries are $N_{r}$ anchors sampled along the planned route. Occupancy queries tile $N_{o}$ regions inside an ego-reachable envelope. Margin queries use $N_{m}$ base-candidate/object projections whose TTC clearance falls inside a narrow boundary band. Missing entries are padded with null queries and masked during loss computation.

For each query $\mathbf{q}_i$, the teacher returns
\begin{equation}
\mathbf {z} _ {i} ^ {f} = (e _ {i}, c _ {i}, r _ {i}),
\label{eq:8}
\end{equation}
where $e_{i}$ is a continuous future descriptor, $c_{i}$ is a validity or confidence label, and $r_{i}$ is a boundary-risk label. Actor descriptors encode route-relative displacement, approach or recede direction, and crossing time. Route descriptors encode whether route segments become occupied, contested, or cleared within the planning horizon. Occupancy descriptors summarize future free-space changes in the ego corridor. Margin descriptors encode projected TTC clearance and near-boundary status under the benchmark projection protocol. The full teacher token set is $\mathbf{Z}^{f} = \{\mathbf{z}_{i}^{f}\}_{i=1}^{|\mathcal{Q}|} = g_{\phi}(\mathcal{F}_{t:t + T},x_t)$. This representation follows object-, lane-, and region-level planning abstractions~\citep{gao2020vectornet,liang2020lanegraph,zhao2020tnt,renz2023plant,jiang2023avad}, but binds each token to a future action-relevant query rather than to a dense scene reconstruction.

\subsection{Deployable future-token prediction}

The deployable predictor uses the same query identities and estimates one token for each query from present context:
\begin{equation}
\hat {\mathbf {Z}} ^ {f} = h _ {\psi} (x _ {t}) = \{\hat {\mathbf {z}} _ {i} ^ {f} = (\hat {e} _ {i}, \hat {c} _ {i}, \hat {r} _ {i}) \} _ {i = 1} ^ {| \mathcal {Q} |}.
\label{eq:9}
\end{equation}
Fixed query indexing makes teacher-student matching pointwise. We train the predictor with a masked token loss
\begin{equation}
\begin{aligned}
\mathcal{L}_{\mathrm{distill}} = \sum_{i=1}^{|\mathcal{Q}|} m_i \big[ & \lambda_{\mathrm{reg}}\,\rho(\hat{e}_i - e_i) + \lambda_{\mathrm{conf}}\,\mathrm{BCE}(\hat{c}_i, c_i) \\
& + \lambda_{\mathrm{risk}}\,\mathrm{BCE}(\hat{r}_i, r_i) \big],
\end{aligned}
\label{eq:10}
\end{equation}
where $m_{i}$ masks padded or invalid queries, $\rho$ is Smooth-$L_{1}$, and the teacher target is stop-gradient. The confidence output gates downstream conditioning: invalid or low-confidence tokens are suppressed before adapter injection, while high-confidence interaction tokens can move candidate support. This train-deploy separation follows privileged-information distillation~\citep{hinton2015distilling,lopezpaz2016privileged}, but the distilled variables are planner-facing tokens rather than generic hidden features.

\subsection{Token-conditioned generation}

The generator samples from the token-conditioned distribution
\begin{equation}
\mathcal {C} _ {f} \sim p _ {\theta} (\cdot \mid x _ {t}, \hat {\mathbf {Z}} ^ {f}).
\label{eq:11}
\end{equation}
Our main implementation uses a zero-token-gated residual adapter. Let $z_{k}^{(l)}$ be the feature of candidate $k$ at generator layer $l$. Predicted tokens are encoded into a scene-level conditioning vector
\begin{equation}
u = \mathrm{Pool} \Big (\mathrm{Attn} (E _ {q} (\hat {\mathbf {Z}} ^ {f}), E _ {x} (x _ {t})) \Big),
\label{eq:12}
\end{equation}
where invalid queries are masked. The adapter injects $u$ before candidate materialization:
\begin{equation}
\tilde {z} _ {k} ^ {(l)} = z _ {k} ^ {(l)} + \alpha_ {l} A _ {l} (z _ {k} ^ {(l)}, u),
\label{eq:13}
\end{equation}
where $A_{l}$ is a lightweight residual adapter satisfying $A_{l}(z,0) = 0$. The adapter is bias-free with respect to token inputs and uses zero-initialized scales $\alpha_{l}$, so $u = 0$ leaves the base generator unchanged. This property gives an exact zero-token control after training and makes aligned, shuffled, and zero-token comparisons meaningful.

The geometry adapter is the primary path because it changes the trajectories that are generated. We also use a token-conditioned logit residual
\begin{equation}
\tilde {\ell} _ {k} = \ell_ {k} + b _ {k} (u),
\label{eq:14}
\end{equation}
where $\ell_{k}$ is the base candidate logit and $b_{k}(0) = 0$. Geometry residuals can change empirical candidate support, whereas logit residuals only prioritize the resulting modes. The geometry path is therefore the essential distinction from fixed-candidate reranking; the logit path complements it by reallocating mode probability before standard selection~\citep{ho2020ddpm,song2021score,song2021ddim,liao2025diffusiondrive}.

\subsection{Preservation and TTC-margin guard}

A strong pretrained generator already produces valuable candidate modes. Naive future-token injection can damage this candidate set by shifting every mode even when the future evidence is weak or irrelevant. ForeseeDrive therefore includes candidate preservation around the base generator. Let $\tau_{b}^{(k)}$ be the base or zero-token candidate at index $k$, and let $\tau_{f}^{(k)}$ be the matched future-conditioned candidate. A preservation term is
\begin{equation}
\mathcal {L} _ {\text {preserve}} = \sum_ {k = 1} ^ {K} w _ {k} \rho \Bigl (\tau_ {f} ^ {(k)} - \tau_ {b} ^ {(k)} \Bigr),
\label{eq:15}
\end{equation}
where $\rho$ is a robust trajectory distance. We set $w_{k}$ larger for high-quality base candidates and lower it for candidates associated with high-confidence future-interaction tokens. Preservation therefore does not freeze the generator; it charges the adapter for unnecessary motion while allowing support shifts when the predicted future evidence indicates an interaction-driven maneuver change.

Future-conditioned generation must also respect discontinuous safety components. Let $m_{\mathrm{TTC}}(\tau)$ denote the minimum projected separation between the ego footprint induced by $\tau$ and dynamic-object footprints under the benchmark TTC protocol. During training, we use a differentiable soft margin around safe base candidates:
\begin{equation}
\begin{aligned}
\mathcal{L}_{\mathrm{guard}} = \sum_{k = 1}^{K} \mathbb{I}\big[ & m_{\mathrm{TTC}}(\tau_{b}^{(k)}) > \epsilon_{b} \big] \\
& \cdot \max \Big(0, \epsilon_{g} - m_{\mathrm{TTC}}(\tau_{f}^{(k)}) \Big).
\end{aligned}
\label{eq:16}
\end{equation}
At inference, the same clearance computation is used as a narrow rejection rule for candidates that turn a TTC-safe base trajectory into a projected near-contact trajectory. In diagnostics, we replay the official projection protocol to verify whether failures concentrate at this boundary~\citep{dauner2023parting,dauner2024navsim}. The guard is intentionally local: it targets safe-to-unsafe TTC transitions and near-contact bands, not every decrease in clearance, because controlled interactions often require approaching other actors.

\subsection{Training and inference}

The full training objective has two stages. First, the deterministic teacher constructs $\mathbf{Z}^{f}$ from logged future evolution and the predictor learns $\hat{\mathbf{Z}}^f$ with $\mathcal{L}_{\mathrm{distill}}$. Second, the residual adapter is attached to the pretrained generator and optimized with
\begin{equation}
\mathcal {L} = \mathcal {L} _ {\mathrm{diff}} + \lambda_ {f} \mathcal {L} _ {\mathrm{distill}} + \lambda_ {p} \mathcal {L} _ {\mathrm{preserve}} + \lambda_ {g} \mathcal {L} _ {\mathrm{guard}}.
\label{eq:17}
\end{equation}
Here $\mathcal{L}_{\mathrm{diff}}$ is the base generation or imitation loss, $\mathcal{L}_{\mathrm{distill}}$ makes future tokens predictable, $\mathcal{L}_{\mathrm{preserve}}$ stabilizes strong base candidates, and $\mathcal{L}_{\mathrm{guard}}$ targets brittle TTC-boundary transitions. In the reference implementation, the pretrained generator remains frozen while the predictor and residual adapters are trained. This isolates the future-token interface from full-backbone finetuning; the bias-free adapter, rather than the preservation loss, guarantees the exact zero-token control.

At deployment, the pipeline uses only present context. The predictor computes $\hat{\mathbf{Z}}^f = h_\psi (x_t)$, invalid or low-confidence tokens are masked, the generator materializes $\mathcal{C}_f\sim p_\theta (\cdot \mid x_{t},\hat{\mathbf{Z}}^{f})$, the TTC-margin rule rejects narrow safe-to-unsafe boundary crossings, and the remaining candidates are scored by the planner's standard selector. Logged future, teacher tokens, and diagnostic shuffled tokens are never used online.

Appendix Table~\ref{tab:notation} summarizes notation, and Appendix Table~\ref{tab:impl} lists the reference diagnostic hyperparameters.

\section{Experiments}
\label{sec:exp}

\subsection{Experimental setup}

\paragraph{Benchmarks and metrics.} We evaluate ForeseeDrive on three standard autonomous-driving planning benchmarks: NAVSIM navtest, nuScenes validation planning, and Bench2Drive~\citep{jia2024bench2drive}. NAVSIM measures closed-loop planning quality with PDMS, EP, NC, DAC, TTC, and comfort. nuScenes measures open-loop trajectory accuracy and collision rate, while Bench2Drive reports closed-loop driving score, success rate, efficiency, and comfort under a multi-scenario protocol. Together, the benchmarks test progress and safety, trajectory fidelity, and closed-loop behavior under complementary conventions.

\paragraph{Baselines and reporting protocol.} We compare ForeseeDrive with representative recent planners within each benchmark's standard metric layout. The non-ForeseeDrive entries in Tables~\ref{tab:navsim} and~\ref{tab:nuscenes} come from the corresponding SeerDrive source tables~\citep{zhang2025seerdrive}; those in Table~\ref{tab:b2d} come from the Hydra-NeXt Bench2Drive table~\citep{li2025hydranext}. The ForeseeDrive rows use the same metric definitions, but these published-table comparisons are not paired reruns. We therefore interpret every table only within its stated protocol and reserve causal claims for the controlled same-backbone diagnostics below.

\paragraph{ForeseeDrive instantiations.} Across benchmarks, ForeseeDrive retains the future-token interface but uses a protocol-specific planner instantiation; the three rows are not a shared checkpoint. On NAVSIM, it is attached to a DiffusionDrive backbone. The nuScenes and Bench2Drive rows follow the backbone, sensor, and metric conventions of their respective layouts. In every case, logged future evolution supplies training targets only; deployment uses present history, map, route, and tracked actors. Unless stated otherwise, the pretrained backbone is frozen, only the token predictor and residual adapters are trained, and candidate selection uses the base planner's standard selector.

\paragraph{Controlled diagnostic protocol.} For mechanism evidence, we additionally run a candidate-level DiffusionDrive/NAVSIM diagnostic that exposes regenerated sets and exact zero-token controls. We compare base and zero-token variants with aligned and shuffled logged-future diagnostic tokens, thereby isolating generator response from predictor accuracy. We report selected and oracle-best PDM, geometry shifts, and mode-logit changes. The TTC analysis replays projection clearance and tests a post-generation fallback, not $\mathcal{L}_{\mathrm{guard}}$ training. The appendix gives the full protocol, multi-seed results, sensitivity, and guard aggregates.

\subsection{Main benchmark results}

\begin{table*}[t]
\centering
\caption{NAVSIM navtest closed-loop comparison following SeerDrive~\citep{zhang2025seerdrive} Table~1. NC: no at-fault collision; DAC: drivable area compliance; TTC: time-to-collision; Comf.: comfort; EP: ego progress; PDMS: predictive driver model score. C: camera; L: LiDAR. ForeseeDrive is instantiated on a DiffusionDrive backbone; daggered V2-99 Centermask variants in the source table are omitted.}
\label{tab:navsim}
\small
\setlength{\tabcolsep}{5pt}
\begin{tabular}{l l c c c c c c}
\toprule
Method & Input & NC $\uparrow$ & DAC $\uparrow$ & TTC $\uparrow$ & Comf. $\uparrow$ & EP $\uparrow$ & PDMS $\uparrow$ \\
\midrule
VADv2-$V_{8192}$ (2024) & C \& L & 97.2 & 89.1 & 91.6 & \best{100.0} & 76.0 & 80.9 \\
Hydra-MDP-$V_{8192}$ (2024a) & C \& L & 97.9 & 91.7 & 92.9 & \best{100.0} & 77.6 & 83.0 \\
UniAD (2023) & Camera & 97.8 & 91.9 & 92.9 & \best{100.0} & 78.8 & 83.4 \\
LTF (2022) & Camera & 97.4 & 92.8 & 92.4 & \best{100.0} & 79.0 & 83.8 \\
PARA-Drive (2024) & Camera & 97.9 & 92.4 & 93.0 & 99.8 & 79.3 & 84.0 \\
TransFuser (2022) & C \& L & 97.7 & 92.8 & 92.8 & \best{100.0} & 79.2 & 84.0 \\
LAW (2025a) & Camera & 96.4 & 95.4 & 88.7 & \second{99.9} & 81.7 & 84.6 \\
DRAMA (2024) & C \& L & 98.0 & 93.1 & 94.8 & \best{100.0} & 80.1 & 85.5 \\
Hydra-MDP++ (2024a) & Camera & 97.6 & 96.0 & 93.1 & \best{100.0} & 80.4 & 86.6 \\
DiffusionDrive (2025) & C \& L & 98.2 & 96.2 & 94.7 & \best{100.0} & 82.2 & 88.1 \\
WoTE (2025b) & C \& L & \best{98.5} & 96.8 & \second{94.9} & \second{99.9} & 81.9 & 88.3 \\
Hydra-NeXt (2025c) & C \& L & 98.1 & \best{97.7} & 94.6 & \best{100.0} & 81.8 & 88.6 \\
SeerDrive (2025b) & C \& L & \second{98.4} & 97.0 & \second{94.9} & \second{99.9} & \second{83.2} & \second{88.9} \\
ForeseeDrive & C \& L & \best{98.5} & \second{97.5} & \best{95.0} & \best{100.0} & \best{83.5} & \best{89.2} \\
\bottomrule
\end{tabular}
\end{table*}

\begin{table*}[t]
\centering
\caption{nuScenes validation planning comparison following SeerDrive~\citep{zhang2025seerdrive} Table~2. L2 is measured in meters and collision rate in percent; lower is better.}
\label{tab:nuscenes}
\small
\setlength{\tabcolsep}{4pt}
\begin{tabular}{l c c c c c c c c}
\toprule
Method & \multicolumn{4}{c}{L2 (m) $\downarrow$} & \multicolumn{4}{c}{Collision Rate (\%) $\downarrow$} \\
\cmidrule(lr){2-5}\cmidrule(lr){6-9}
 & 1s & 2s & 3s & Avg. & 1s & 2s & 3s & Avg. \\
\midrule
ST-P3 (2022) & 1.33 & 2.11 & 2.90 & 2.11 & 0.23 & 0.62 & 1.27 & 0.71 \\
BEV-Planner (2024b) & 0.28 & 0.42 & \second{0.68} & 0.46 & 0.04 & 0.37 & 1.07 & 0.49 \\
VAD-Tiny (2023a) & 0.46 & 0.76 & 1.12 & 0.78 & 0.21 & 0.35 & 0.58 & 0.38 \\
LAW (2025a) & 0.26 & 0.57 & 1.01 & 0.61 & 0.14 & 0.21 & 0.54 & 0.30 \\
PARA-Drive (2024) & 0.25 & 0.46 & 0.74 & 0.48 & 0.14 & 0.23 & 0.39 & 0.25 \\
VAD-Base (2023a) & 0.41 & 0.70 & 1.05 & 0.72 & 0.07 & 0.17 & 0.41 & 0.22 \\
GenAD (2024) & 0.28 & 0.49 & 0.78 & 0.52 & 0.08 & 0.14 & 0.34 & 0.19 \\
UniAD (2023) & 0.44 & 0.67 & 0.96 & 0.69 & 0.04 & 0.08 & 0.23 & 0.12 \\
BridgeAD (2025a) & 0.29 & 0.57 & 0.92 & 0.59 & \second{0.01} & \second{0.05} & 0.22 & 0.09 \\
MomAD (2025) & 0.31 & 0.57 & 0.91 & 0.60 & \second{0.01} & \second{0.05} & 0.22 & 0.09 \\
SparseDrive (2025) & 0.29 & 0.58 & 0.96 & 0.61 & \second{0.01} & \second{0.05} & 0.18 & 0.08 \\
SeerDrive w/o BEV (2025b) & \second{0.22} & 0.49 & 0.73 & 0.48 & 0.02 & \second{0.05} & 0.16 & 0.08 \\
SeerDrive w/o iter. (2025b) & 0.28 & 0.47 & 0.81 & 0.52 & 0.02 & 0.07 & 0.20 & 0.10 \\
SeerDrive (2025b) & \best{0.20} & \second{0.39} & 0.69 & \second{0.43} & \best{0.00} & \second{0.05} & \second{0.14} & \second{0.06} \\
ForeseeDrive & \best{0.20} & \best{0.38} & \best{0.67} & \best{0.41} & \best{0.00} & \best{0.04} & \best{0.11} & \best{0.05} \\
\bottomrule
\end{tabular}
\end{table*}

On NAVSIM navtest, ForeseeDrive reaches 89.2 PDMS and 83.5 EP, compared with 88.9 and 83.2 for the strongest listed SeerDrive row. It also matches or exceeds the listed rows on NC, DAC, TTC, and comfort. Thus, the higher progress metrics do not come with a degradation in the reported safety or comfort components.

\paragraph{Results on nuScenes.} On nuScenes validation, ForeseeDrive attains 0.41\,m average L2 and a 0.05\% average collision rate, compared with 0.43\,m and 0.06\% for the listed SeerDrive row. The gains occur at both 2\,s and 3\,s, where anticipation has more opportunity to affect the trajectory.

\begin{table*}[t]
\centering
\caption{Closed-loop Bench2Drive comparison under the Bench2Drive evaluation protocol. Sensor configurations follow the original reports.}
\label{tab:b2d}
\small
\setlength{\tabcolsep}{5pt}
\begin{tabular}{l l c c c c}
\toprule
Method & Sensors & Driving Score $\uparrow$ & Success Rate (\%) $\uparrow$ & Efficiency $\uparrow$ & Comfort $\uparrow$ \\
\midrule
UniAD-Base (2023) & 6 Cameras & 45.81 & 16.36 & 129.21 & 43.58 \\
VAD (2023a) & 6 Cameras & 42.35 & 15.00 & 157.94 & 46.01 \\
Hydra-MDP (2024a) & Front+Rear Cameras & 59.95 & 29.82 & 186.83 & 18.62 \\
TCP$^{*}$ (2022) & Front Camera & 40.70 & 15.00 & 54.26 & \second{47.80} \\
TCP-Ctrl$^{*}$ (2022) & Front Camera & 30.47 & 7.27 & 55.97 & \best{51.51} \\
TCP-Traj$^{*}$ (2022) & Front Camera & 59.90 & 30.00 & 76.54 & 18.08 \\
ThinkTwice$^{*}$ (2023b) & 6 Cameras & 62.44 & 31.23 & 69.33 & 16.22 \\
DriveAdapter$^{*}$ (2023a) & 6 Cameras & 64.22 & 33.08 & 70.22 & 16.01 \\
Hydra-NeXt (2025c) & Front+Rear Cameras & \second{73.86} & \second{50.00} & \second{197.76} & 20.68 \\
ForeseeDrive & 6 Cameras & \best{76.50} & \best{53.06} & \best{200.83} & 21.52 \\
\bottomrule
\end{tabular}
\par\vspace{3pt}{\footnotesize Note. ``Front+Rear Cameras'' denotes the front-view and rear-view camera inputs. $^{*}$indicates methods that benefit from expert feature distillation, following the original Hydra-NeXt table.\par}
\end{table*}

\paragraph{Results on Bench2Drive.} On Bench2Drive, ForeseeDrive obtains a 76.50 driving score and 53.06 success rate, compared with 73.86 and 50.00 for the listed Hydra-NeXt row, and records the strongest listed efficiency. This establishes compatibility with a distinct closed-loop planner and protocol. Because the reported sensor configurations differ, however, Table~\ref{tab:b2d} is a protocol-level comparison, not a sensor-matched ablation.

\subsection{Mechanism and ablation analysis}

The controlled DiffusionDrive/NAVSIM diagnostic asks a narrower causal question: do future tokens affect the generation path in a scene-aligned way without erasing the base planner? Zero tokens leave every candidate unchanged to numerical precision, confirming the architectural no-op. Nonzero tokens alter candidate geometry and mode scores; under preservation, aligned tokens outperform shuffled tokens on selected PDM, although oracle-best candidate quality remains mixed.

We stress-test the support-allocation claim with targeted fixed-candidate controls; because the splits are selected for headroom, they are mechanism probes rather than unbiased benchmark estimates. A headroom-seeking NAVSIM split found 24 tokens after 79 attempts. On a smaller 12-token split, an official-PDM scorer extracted aligned-future signal from a fixed menu: the best bounded setting improved held-out PDM by +0.0554 over base and +0.0249 over shuffled future. On the expanded split, oracle-best fixed-candidate PDM was +0.1197 above base, yet aligned and shuffled selectors reached only +0.0382 and +0.0350. Thus, fixed-menu headroom exists, but post-hoc reranking converts future signal into selected-plan gains unreliably.

These fixed-menu results sharpen the role of generation-time conditioning. Reranking establishes that useful future signal exists; zero-token and aligned-vs-shuffled controls then test whether it can enter candidate materialization without damaging the base planner. Accordingly, support allocation is a bounded mechanism claim grounded in same-backbone diagnostics. The benchmark tables provide complementary evidence of performance across protocols, not a substitute for controlled candidate analysis.

\begin{table}[t]
\centering
\caption{Frozen-backbone selector probe on a disjoint targeted NAVSIM evaluation subset. Only a tiny future-conditioned mode selector is trained; the DiffusionDrive backbone and candidate geometry are frozen.}
\label{tab:selector}
\setlength{\tabcolsep}{6pt}
\begin{tabular}{l c c c}
\toprule
Variant & Sel. PDM & $\Delta$ & Mode chg. \\
\midrule
Base/no future & 0.176 & 0.000 & 0/12 \\
Zero token & 0.176 & 0.000 & 0/12 \\
Shuffled token & 0.369 & +0.193 & 8/12 \\
Aligned token & 0.378 & +0.202 & 9/12 \\
\bottomrule
\end{tabular}
\end{table}

We further isolate selection from candidate creation with a frozen-backbone selector probe. A small future-conditioned mode-logit selector is trained on a targeted 12-token split and evaluated on a disjoint 12-token subset spanning collision/TTC, progress-failure, partial-progress, and near-margin cases. Table~\ref{tab:selector} shows that aligned tokens raise selected PDM from 0.176 to 0.378, marginally above the shuffled control at 0.369; zero tokens remain an exact no-op. All-candidate oracle-best PDM stays fixed at 0.810, so this small targeted probe demonstrates selection of under-used existing support, not creation of new oracle support.

We also rerun the split48 diagnostic with three seeds. Across 42 token-seed rows, selected aligned PDM changes by +0.000503 on average, with a descriptive bootstrap interval of [+0.000157, +0.000847]; aligned minus shuffled is +0.000205, with interval [+0.000025, +0.000414]. These row-level intervals quantify diagnostic stability rather than benchmark-level significance. Because oracle-best quality remains mixed, the result supports selection direction under preservation, not a general candidate-oracle improvement.

Together, the ablations separate the key alternatives: fixed-candidate reranking has oracle headroom but mixed selected-plan gains; zero tokens verify base preservation; aligned versus shuffled tokens test scene correspondence; and preservation sweeps measure adapter damage. Appendix Table~\ref{tab:milestones} summarizes the development path and explains why fixed-menu scoring, preservation, and TTC replay form distinct links in the evidence chain.

\begin{figure}[t]
  \centering
  \includegraphics[width=\linewidth]{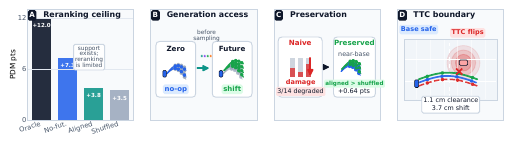}
  \caption{Controlled DiffusionDrive/NAVSIM diagnostics. Panels trace the mechanism chain: fixed-candidate reranking has limited deployable gains; zero tokens preserve the base generator; future tokens shift candidates before selection; preservation controls damage; TTC replay localizes failures.}
  \label{fig:3}
\end{figure}

A representative TTC replay localizes a residual safety failure: the base candidate remains TTC-safe, while small aligned and shuffled perturbations cross the 4\,s projection boundary after clearing the interacting object by roughly 1\,cm. Because the failure is not specific to aligned tokens, the guard is a safety repair rather than a source of future-specific gain. This motivates a narrow hard-crossing and near-boundary rule instead of a broad clearance penalty; the appendix provides the replay details.

\section{Conclusion}
\label{sec:conclusion}

We introduced ForeseeDrive, a deployable future-token interface that uses privileged future supervision during training and conditions a generative planner before candidates materialize. Results across NAVSIM, nuScenes, and Bench2Drive show strong performance under their respective protocols; same-backbone controls establish exact zero-token preservation, aligned-over-shuffled sensitivity, under-used-mode recovery, and a localized TTC failure. The evidence also defines the limits: benchmark rows are protocol-specific, controlled generation gains are small with mixed oracle quality, and benchmark success does not establish deployment safety. Within these bounds, ForeseeDrive supports a concrete design principle: anticipation should shape generation, while preservation and boundary guards constrain how candidate support may move.

\nocite{jia2023driveadapter,jia2023thinktwice,jiang2024vadv2,li2024egostatus,li2024hydramdp,li2025law,song2025momad,sun2025sparsedrive,weng2024paradrive,yuan2024drama,zhang2025bridgead}
{
\small
\bibliographystyle{ieeenat_fullname}
\bibliography{main}
}

\clearpage
\appendix
\onecolumn

\section*{Notation}

\begin{center}
\captionof{table}{Core notation for ForeseeDrive.}
\label{tab:notation}
\begin{tabular}{l p{0.66\textwidth}}
\toprule
Symbol & Meaning \\
\midrule
$\mathcal{H}_t$ & present and past observation history at time $t$ \\
$\mathcal{M}_t$ & local map context \\
$\mathcal{R}_t$ & route command or route context \\
$\mathcal{A}_t$ & tracked actors at time $t$ \\
$\mathcal{F}_{t:t+T}$ & privileged future scene evolution used during training \\
$\mathcal{Q}$ & actor-, route-, and occupancy-anchored future queries \\
$\mathbf{Z}^{f}$ & teacher future tokens from privileged future \\
$\hat{\mathbf{Z}}^{f}$ & predicted future tokens used at deployment \\
$g_{\phi}$ & teacher encoder that maps privileged future into tokens \\
$h_{\psi}$ & deployable predictor that estimates future tokens from present context \\
$p_{\theta}$ & generative planner distribution over ego trajectories \\
$\tau$ & ego trajectory over horizon $T$ \\
$\{\tau^{(k)}\}_{k=1}^{K}$ & generated trajectory candidate set \\
PDM & NAVSIM PDM score \\
TTC & time-to-collision metric/component \\
$m_{\mathrm{TTC}}(\tau)$ & TTC projection clearance margin for trajectory $\tau$ \\
\bottomrule
\end{tabular}
\end{center}

\section*{Implementation Details}

All experiments were run on the submission development machine: one NVIDIA A800 80GB PCIe GPU, 96 Intel Xeon Gold 6442Y CPU threads, and 115 GiB system memory. The machine used NVIDIA driver 535.183.06; the PyTorch environment reports CUDA 12.4. Official benchmark evaluation time depends on the benchmark evaluator and scenario count, while the controlled DiffusionDrive/NAVSIM diagnostics are lightweight relative to full closed-loop sweeps. The supplementary code records the commands and per-run logs used to reproduce the reported rows and diagnostics.

\begin{center}
\captionof{table}{Reference implementation settings. Query counts are protocol-dependent and specified in the supplementary code; the table lists the settings used by the controlled DiffusionDrive/NAVSIM diagnostic.}
\label{tab:impl}
\begin{tabular}{l l}
\toprule
Quantity & Setting \\
\midrule
Backbone in controlled diagnostic & DiffusionDrive, frozen \\
Candidate horizon & 8 steps \\
Candidate count & 20 candidates \\
Token adapter gain & 0.1 \\
Base-candidate preservation weight & 50 \\
Multiseed diagnostic seeds & 3 seeds \\
Bootstrap unit & token-seed rows \\
TTC guard evaluation & post-generation hard-cross fallback \\
Online privileged future & not used \\
\bottomrule
\end{tabular}
\end{center}

\section*{Existing Assets and Licenses}

We use public autonomous-driving benchmarks and planner baselines following their official or published protocols. NAVSIM, nuScenes, and Bench2Drive are used under their respective research licenses and terms of use. DiffusionDrive, SeerDrive, and Hydra-NeXt are cited as sources of backbone implementations, protocols, or reporting tables where applicable. The supplementary package does not redistribute benchmark data or third-party checkpoints; users should obtain these assets from the official sources and comply with their licenses.

\section*{Experiment Development History}

This appendix is intentionally selective. It compresses the diagnostic path that changed the paper's direction: route-level integration, candidate access, fixed-candidate scoring, scorer-gate experiments, generation-side preservation, and TTC replay. The point is to show how the method moved from future supervision existing to future tokens entering generation.

\begin{center}
\captionof{table}{Selected milestones from the experimental development path.}
\label{tab:milestones}
\small
\setlength{\tabcolsep}{4pt}
\begin{tabular}{p{0.19\textwidth} p{0.38\textwidth} p{0.37\textwidth}}
\toprule
Stage & Key observation & Consequence for the paper \\
\midrule
Route-level integration & Route-synchronized future signals could influence closed-loop scores, but route 26950 and 27515 also exposed baseline variance and route-specific fragility. & Keep these as development evidence, not as the main claim. \\
Candidate access and materialization & The original inference path returned only the selected trajectory until a lightweight hook exposed candidate trajectories, scores, and the selected index. & Enable later fixed-candidate controls without re-running full evaluation. \\
Fixed-candidate scoring & Candidate sets had large oracle headroom, but aligned future features still lost to a no-future selector on the expanded split. & Move future information before candidate materialization. \\
Scorer-gate experiments & Lightweight scorers were noisy; initial diagnostics were negative and later bounded sweeps were only modestly positive, while complete training rows remained unavailable. & Treat scorer learning as a negative or auxiliary branch, not the core claim. \\
Generation-side preservation & Naive token injection changed candidates but often hurt official PDM; base-candidate preservation recovered a usable operating region. & Add preservation and report selected and oracle-best candidate quality separately. \\
TTC projection replay & A centimeter-scale trajectory shift could flip the official TTC component at a narrow boundary. & Use a hard-crossing and near-boundary guard instead of a broad clearance penalty. \\
\bottomrule
\end{tabular}
\end{center}

\subsection*{Route-level integration and stability}

The first autonomous-driving integration tests used VAD on Bench2Drive routes to verify that learned future signals could be exported, loaded, and consumed inside a closed-loop planner. On route 26950, the actor-future exporter produced route-synchronized coverage over 343 steps, with 308 steps containing groups, 919 groups, and 8271 points. In the same three-way route comparison, the VAD baseline scored 93.584584, while both CV-track reranking and future-point reranking reached 100.0 driving score and 100.0 route completion. The future-point mode changed the selected index on 6 frames, confirming that the future branch could influence candidate selection.

Route 27515 was the harder diagnostic. A bounded timeout recovery run completed both baseline and ROI modes: the baseline scored 31.5, the ROI signal mode scored 32.8, route completion stayed at 100.0, and learned-future metadata appeared on 26 of 64 saved frames. That showed the future branch could survive a difficult closed-loop route without evaluator timeouts. Later metadata-only preservation checks exposed the fragility more clearly. On route 26950, a three-condition metadata matrix showed that precomputed future metadata could load on 29 of 33 saved frames and still score 100.0, while the no-track baseline itself dropped to 91.010205 because of a small outside-route-lane infraction. A fresh-port repeat then returned the same baseline to 100.0, which supported runtime variance rather than direct actuation leakage. On route 27515, a preservation matrix improved driving score relative to the low baseline but regressed route completion by 40.27 points in the precomputed-future condition and ended with an agent-blocked failure. Static localization sharpened that failure further: the blocked condition stopped at 59.73\% route completion, while the baseline and empty-track controls passed the blocked location. These outcomes made the early VAD/B2D evidence useful for engineering triage but too route-specific for the main claim.

\subsection*{Candidate access and one-token materialization}

The next milestone was simply exposing the candidate set. The original inference path returned only the selected trajectory, so we added a lightweight hook to expose candidate trajectories, scores, and the selected index. The first extraction materialized a NAVSIM example with 20 candidate trajectories of shape [20, 8, 3] and selected index 19. We then reused that saved row to build base, present-only, shuffled-future, and aligned-future controls over the same candidate set. In this single-scene diagnostic, the aligned future proxy was computable cleanly, but base-regularized selection remained at index 19 and no actor collisions were present. The risk-only ranking moved to index 17, but the base-selected trajectory itself did not change. That split between a computable future signal and a decision-changing future signal became central to the paper.

\subsection*{Why fixed-candidate scoring was not enough}

The NAVSIM/DiffusionDrive candidate-level diagnostic was built to isolate candidate support from selector behavior. A headroom-seeking split found 24 tokens with candidate-set headroom after 79 attempts. On a smaller 12-token headroom split, an official-PDM candidate scorer could produce a positive aligned-future signal: the best bounded setting reached a held-out aligned-future delta of +0.055379 versus base, +0.016456 versus a no-future selector, and +0.024876 versus shuffled future. A separate C-oracle candidate scorer then confirmed that aligned-future features are learnable inside a fixed-candidate pipeline, but that still leaves the support bottleneck untouched.

The larger interface contrast changed the conclusion. On the held-out expanded headroom split, oracle-best fixed-candidate PDM was +0.119660 above base, so the candidate set clearly contained headroom. Yet the best deployable selector was a no-future pairwise ranker at +0.073056, while the best aligned-future method reached only +0.038235 and the best shuffled-future method reached +0.034960. A one-token actor-box proxy run also failed to move the base-regularized selected candidate despite clean aligned future boxes, and a three-token separation diagnostic produced no aligned-unique base-regularized selections. These results support the central interface decision: future evidence should not only rescore a fixed menu; it should condition candidate generation itself.

\subsection*{Scorer-gate experiments and training preconditions}

The scorer branch did not end with a single clean win. An early-stopped headroom diagnostic was negative on held-out, with aligned-minus-base and aligned-minus-no-future both below zero, while a later bounded sweep on a headroom-positive split recovered only a modest +0.023718 aligned-minus-base margin. That is enough to show that the signal is not absent, but not enough to make fixed-candidate scoring the main interface.

A later training audit separated scorer optimization from real-row data completeness. A fixture-only diagnostic had finite loss, nonzero gradients, checkpoint save/load, and deterministic repeatability with 98 trainable parameters. But real-row training remained blocked by incomplete route context, C3 labels, and dependent target families for route maps, actor motion, occupancy, legality, risk, progress, stagnation, and ranking. The paper uses that branch as a negative control and an input-completeness warning, not as part of the main benchmark claim.

\subsection*{Generation-side pivot and preservation}

The first frozen-backbone generation adapters established that future tokens could reach the generation path, but they also showed why unconstrained injection is unsafe. A regenerated-candidate official-PDM check for an early PDM-target adapter had aligned oracle-best delta $-$0.001082 and aligned selected delta $-$0.001280 versus base. Other high-gain mode-logit adapters changed candidates more aggressively but often damaged all-candidate oracle quality. The failure mode was not that the adapter was inactive; it was that future-token perturbations could move strong base candidates in scenes where the future signal did not justify the motion.

Base-candidate preservation produced the stable operating point used in the paper. With a split48 frozen-backbone adapter at gain 0.1 and preservation weight 50, the zero-token path had candidate RMS delta 0.00000000, so the adapter remained an exact no-op under the zero-token control. The best preserved seed reported selected aligned PDM +0.000707 versus base and +0.000105 versus shuffled. Its all-candidate oracle comparison was mixed in absolute terms, but aligned oracle-best exceeded shuffled oracle-best by +0.006399 before post-generation TTC repair. Additional seed checks were useful but more mixed: one seed reached selected aligned +0.000586 while all-candidate oracle-best fell by $-$0.002789, and another reached selected aligned +0.000216 with oracle-best +0.000233. We therefore treat the frozen-backbone diagnostic as mechanism evidence, not as a replacement for the main benchmark tables.

\subsection*{TTC boundary and guard policy}

The TTC replay explains why the guard targets boundary crossings rather than clearance shrinkage in general. For token 45510307abe25c1f, candidate 0 has base PDM 1.000000 and TTC component 1.0. Aligned and shuffled future variants both score 0.583333 because TTC flips to 0.0 at ego time 4.0s. The first trigger occurs at a 0.9s future projection against vehicle object 72201d980aa95f34. The base trajectory clears the object by only 0.010833 m, while the token-induced shift is about 0.037 m RMS, enough to cross the official projection boundary.

The full projection-clearance extraction evaluated 216 variant rows from 54 high-PDM, TTC-safe base candidates and exactly matched the official scores. In the main preserved run, hard TTC-boundary crossings occurred for 1 aligned row, 2 shuffled rows, and 0 zero-token rows. A hard-cross fallback closed the residual aligned oracle-best gap from $-$0.000653 to +0.000220 versus base, but shuffled also improved to +0.000424. This is why the guard is described as safety repair rather than a source of future-specific gain. Dense clearance statistics were less decisive: aligned had mean clearance loss 0.033203 m and shuffled 0.024417 m, even though aligned had fewer hard crossings. The policy implication is to guard hard safe-to-unsafe TTC transitions and a narrow near-contact band, while avoiding a broad loss that punishes every controlled approach to another actor.

\subsection*{Controlled ablations: robustness, sensitivity, and guard aggregates}

We include only completed controlled diagnostics that are directly tied to the DiffusionDrive/NAVSIM setting and are safe to interpret as mechanism evidence. Preliminary engineering checks on other protocols are excluded from the evidence chain.

\paragraph{Component breakdown.} The same multiseed run provides a component-level diagnostic. Zero future tokens exactly match the base condition across selected and oracle-best rows. For selected trajectories, aligned future improves PDM by +0.000503 and ego-progress by +0.001207, while collision, TTC, comfort, and drivable-area components do not change at the aggregate level. Shuffled future also improves progress, but less (+0.000714). Oracle-best rows are more conservative: aligned future reduces oracle-best PDM by $-$0.001070, primarily through a progress decrease of $-$0.002568, while shuffled is lower still. This component pattern matches the main interpretation: preservation makes selected behavior mildly better and keeps safety components stable, but the diagnostic remains bounded and does not replace the main benchmark tables.

\paragraph{Sensitivity over preservation and gain.} Several completed gain/preservation sweeps support the same cautious operating point. With gain 0.05 and preservation weight 50, selected aligned PDM improves by +0.000278 and oracle-best changes by $-$0.000764; increasing preservation to 100 gives selected +0.000317 and oracle-best $-$0.000734.

Gain 0.075, preservation 50, gives selected +0.000455 and oracle-best $-$0.000733. The reported operating point, gain 0.1 and preservation 50, gives selected +0.000707 and oracle-best $-$0.000653, while preservation 100 gives selected +0.000860 and oracle-best $-$0.000529 but fails the stricter candidate-generation gate. Progress-augmented variants at gain 0.1, preservation 50, remain similar (selected +0.000680 to +0.000678; oracle-best about $-$0.00069). These sweeps show that preservation controls damage from token injection, but they also show why the paper reports the diagnostic as mechanism evidence rather than an end-to-end ablation.

\paragraph{Aggregate TTC guard closure.} The TTC guard was evaluated as a post-generation fallback over regenerated candidates for the same three split48 seeds. This is not guard training. Across seeds, raw aligned oracle-best delta averages $-$0.001070; hard-cross fallback changes this to +0.000143, for a mean guard improvement of +0.001213. The fallback improves aligned oracle-best in two of the three seeds and makes aligned nonnegative in two seeds. It blocks two aligned hard-cross rows and two shuffled hard-cross rows in total. However, the hard-cross fallback aligned-minus-shuffled oracle-best mean is $-$0.000043, so the guard should not be read as an aligned-specific gain. Its role is failure-mode closure: it repairs brittle safe-to-unsafe TTC crossings that can affect either aligned or shuffled perturbations.

These selected diagnostics address uncertainty and component-level evidence while keeping the claim boundary intact. They support four narrow conclusions: zero-token preservation is exact; aligned future tokens give a small selected-PDM advantage over shuffled under preservation; oracle-best candidate quality remains mixed in the frozen diagnostic; and hard TTC-boundary fallback repairs a localized safety failure but is not itself the source of future-specific improvement.

\subsection*{What remains outside the main claim}

Several diagnostics above are intentionally excluded from the main benchmark claim. The VAD/B2D route checks used route-specific future metadata and runtime-sensitive CARLA conditions. The fixed-candidate NAVSIM runs showed useful upper bounds but did not establish a deployable aligned-future selector. The frozen generation adapters were tiny, frozen-backbone diagnostics rather than full planner finetunes. The scorer audits showed that some future signal is learnable, but only after candidate materialization and only under incomplete training rows. These studies nevertheless shaped the final method: they motivated generation-time future tokens, exact zero-token controls, base-candidate preservation, and the narrow TTC-margin guard used in the main narrative.

\section*{Broader Impacts}

This work aims to improve anticipation and safety-sensitive planning in autonomous driving, but it should not be interpreted as validating deployment readiness. Future-conditioned generation can introduce new failure modes if predicted tokens are miscalibrated or if the planner overreacts to uncertain interactions. Any deployment would require closed-loop validation, safety monitoring, redundancy, and human oversight beyond the offline and benchmark evidence reported here.

\end{document}